\documentclass{article}

\usepackage{amssymb}
\usepackage{amsmath}
\usepackage{amsthm}
\usepackage[a4paper]{geometry}
\usepackage{multirow}

\usepackage{algorithm}
\usepackage{algorithmic}
\usepackage[numbers]{natbib}

\usepackage{authblk}       

\usepackage{subfig}
\usepackage{graphicx}

\usepackage{tabularx}

\usepackage{booktabs}
\usepackage[table]{xcolor}

\definecolor{cvprblue}{rgb}{0.21,0.49,0.74}
\usepackage{hyperref}

\hypersetup{
  colorlinks=true,
  linkcolor=cvprblue,      
  citecolor=cvprblue,      
  urlcolor=magenta          
}

\usepackage{enumitem}
\usepackage{wrapfig}

\title{VisualToolAgent (VisTA): A Reinforcement Learning Framework for Visual Tool Selection}

\author{
Zeyi Huang$^{1}$,\quad Yuyang Ji,\quad Anirudh Sundara Rajan$^{1}$,\quad Zefan Cai$^{1}$,\quad \vspace{-0.6em} \\ Wen Xiao$^{2}$,\quad 
Haohan Wang$^{3}$,\quad Junjie Hu$^{1}$,\quad Yong Jae Lee$^{1}$\\ \vspace{1.2em} 
{$^1$University of Wisconsin-Madison \hspace{1.0cm} $^2$Microsoft} \hspace{1.0cm} $^3$UIUC\\
}
\date{}

\begin{document}
\maketitle

\begin{abstract}
We introduce \textbf{VisTA}, a new reinforcement learning framework that empowers visual agents to dynamically explore, select, and combine tools from a diverse library based on empirical performance. Existing methods for tool-augmented reasoning either rely on training-free prompting or large-scale fine-tuning; both lack active tool exploration and typically assume limited tool diversity, and fine-tuning methods additionally demand extensive human supervision. In contrast, VisTA leverages end-to-end reinforcement learning to iteratively refine sophisticated, query-specific tool selection strategies, using task outcomes as feedback signals. Through Group Relative Policy Optimization (GRPO)~\cite{shao2024deepseekmath}, our framework enables an agent to autonomously discover effective tool-selection pathways without requiring explicit reasoning supervision. Experiments on the ChartQA, Geometry3K, and BlindTest benchmarks demonstrate that VisTA achieves substantial performance gains over training-free baselines, especially on out-of-distribution examples. These results highlight VisTA's ability to enhance generalization, adaptively utilize diverse tools, and pave the way for flexible, experience-driven visual reasoning systems. Project website: \href{https://oodbag.github.io/vista_web/}{https://oodbag.github.io/vista\_web/}.

\end{abstract}

\section{Introduction}
\label{sec:intro}

Recent advances in Large Language Models (LLMs)~\cite{brown2020language, touvron2023llama, achiam2023gpt} and Vision Language Models (VLMs)~\cite{liu2023visual, yang2024qwen2, hurst2024gpt} have unlocked impressive capabilities across tasks such as mathematical problem solving, code generation, and visual question-answering. However, these models are still inherently limited by the static nature of their architectures and the fixed information stored in their weights.
To overcome these constraints, recent work explores augmenting LLMs and VLMs with external tools~\cite{chen2022program, gou2023tora, gao2023pal, patil2024gorilla, hu2024visual, gupta2023visual, suris2023vipergpt}, dramatically expanding their functionality. Tool augmentation enables access to expert knowledge sources and dynamic computation, such as invoking a Python interpreter for self-verification, thereby enhancing reasoning performance on complex tasks.

However, the current paradigm for tool integration faces significant limitations in both LLMs and VLMs. Current approaches typically either rely on large-scale fine-tuning with human supervision to teach LLMs how to invoke tools~\cite{schick2023toolformer, liu2024llava} or depend purely on the LLMs' internal world knowledge in a training-free manner~\cite{gou2023tora, lu2025octotools, hu2024visual}. 
These methods often rely on tool demonstrations~\cite{schick2023toolformer, liu2024llava} or detailed tool descriptions to instruct LLMs on their usage~\cite{hu2024visual, lu2025octotools}. As a result, they lack the ability to automatically explore, select, or adapt tool choices based on the specific characteristics of each query, particularly when multiple tools of the same type with varying capabilities are available which is common in real world settings.
The challenge is particularly pronounced when integrating tools with unknown, partially documented capabilities or inconsistently performing capabilities, where actual performance may differ from descriptions. When retrieving tools from diverse sources, the LLMs lack comprehensive knowledge of their strengths and weaknesses. Without a mechanism for experiential learning, the system cannot determine optimal tool selection or discover synergistic tool combinations that might emerge through collaborative deployment.

\begin{figure}[t]
\centering
\includegraphics[width=0.9\textwidth]{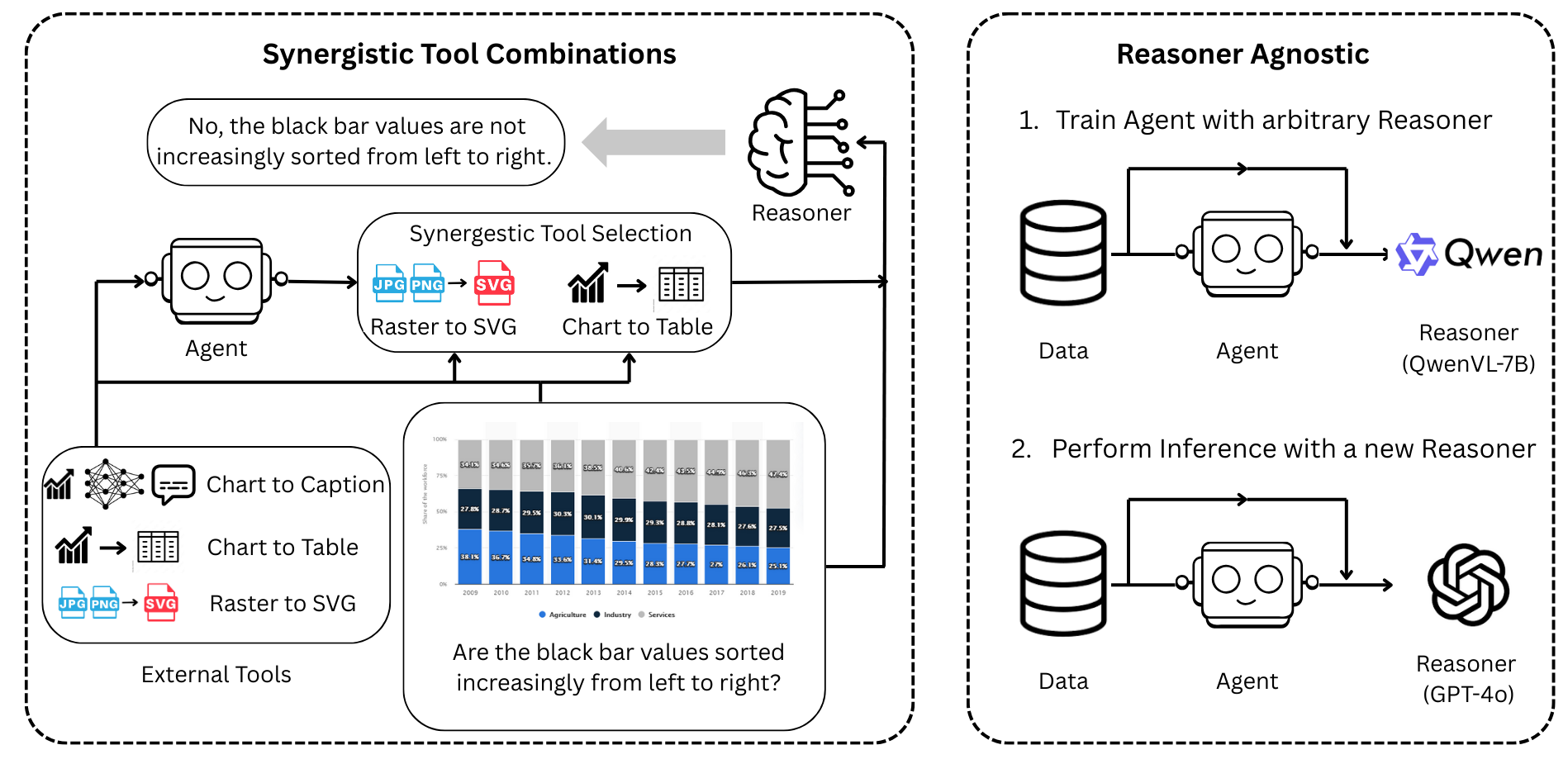}
\caption{\textbf{Overview of VisTA}. (Left) Our method trains an agent to autonomously discover effective combinations of visual tools without human supervision. (Right) By decoupling the agent from the reasoner, the learned policy can be seamlessly integrated with a wide range of reasoning models.}
\label{fig:teaser}
\end{figure}

In realistic applications, tools vary substantially in functionality and applicability across different problem domains. Within each tool category, individual implementations exhibit varying capabilities that make them differentially effective across contexts. This presents a sophisticated decision-making challenge ideally suited for reinforcement learning (RL)~\cite{sutton1998reinforcement}.
RL's intrinsic exploration-exploitation mechanism enables agents to systematically assess and adaptively identify the most effective tools based on empirical performance rather than pre-specified rules. Through iterative interactions with its environment, an RL agent can learn adaptive strategies that dynamically adjust tool combination based on specific queries, or even potentially discover non-obvious tool utility patterns that may not be apparent from tool descriptions alone.

Therefore, we introduce a new RL framework, VisualToolAgent (VisTA), that trains autonomous agents to intelligently select optimal tools from multiple available options. Unlike training-free~\cite{gou2023tora, lu2025octotools, hu2024visual} and fine-tuning approaches~\cite{schick2023toolformer, liu2024llava}, our RL-based method inherently supports exploration-exploitation mechanisms, allowing agents to systematically experiment with various tool combinations through iterative interactions.

In this work, we focus on the visual reasoning task. Our framework consists of an autonomous agent that learns through end-to-end RL training to dynamically select optimal tools for guiding a fixed VLM in solving complex visual reasoning problems. As a bonus, our framework allows the VLM itself to remain frozen during RL training, which means that the agent's learned selection strategies can be transferred to different reasoning models without retraining, a critical advantage for deployment flexibility.  Our framework employs the Group Relative Policy Optimization (GRPO)~\cite{shao2024deepseekmath} algorithm to enable our agent to autonomously discover effective tool-selection pathways entirely from scratch, without explicit reasoning examples. For a detailed look at how the agent performs inference and selects tools in practice, see the examples in Fig \ref{fig:geom_viz}, \ref{fig:chart_viz}.

We evaluate our method on ChartQA~\cite{masry2022chartqa}, a challenging benchmark for visual reasoning that requires models to interpret numerical data, textual labels, and complex visual structures, such as precisely estimating bar heights in charts. We also test on Geometry3K~\cite{lu2021inter}, which assesses fine-grained diagram understanding and logical reasoning. This task requires models to accurately parse visual elements (e.g., figures, labels) and align them with textual question conditions to perform math-based reasoning.
Our experimental results demonstrate that our RL-based approach significantly outperforms training-free methods. The performance gap widens further when testing on Out-of-Distribution (OoD) variants, a robust version that introduces perturbations to the standard dataset. This demonstrates our method's superior ability to generalize to novel scenarios and maintain performance under challenging visual conditions.

\section{Related Work}

\textbf{Tool-Augmented Reasoning.}  LLMs have shown significantly improved reasoning capabilities when augmented with external tools such as search engines~\cite{komeili2021internet}, calculators~\cite{cobbe2021training}, and Python interpreters~\cite{chen2022program, gao2023pal}. Programming-based approaches~\cite{chen2022program, gao2023pal}, for example, integrate Python interpreters to simplify intermediate steps and validate final outputs, enhancing accuracy on mathematical tasks. 
Similar strategies have been adopted in the visual domain. Recent VLM methods~\cite{gupta2023visual, suris2023vipergpt, hu2024visual} generate Python code to invoke specialized vision modules, decomposing complex visual tasks into simpler sub-tasks, each addressable by dedicated vision tools. 
However, existing approaches often rely on human demonstrations or annotations~\cite{schick2023toolformer, liu2024llava}, or operate in a training-free manner using only the model’s internal knowledge~\cite{lu2025octotools, hu2024visual}. These methods typically offer limited tool diversity and depend heavily on explicit tool descriptions~\cite{schick2023toolformer, hu2024visual, liu2024llava, lu2025octotools}, lacking the capacity to autonomously explore or adapt tool use to specific queries.
In contrast, our proposed VisTA framework enables VLMs to autonomously explore and select tools based on empirical performance, without human-designed priors. By training an agent with RL, VisTA discovers context-dependent tool selection policies that adaptively tailor tool usage to the nuanced requirements of each visual reasoning task.

\textbf{Reinforcement Learning for Enhanced Reasoning.} RL has shown strong potential in enhancing complex reasoning abilities and enabling inference scaling. Models like OpenAI’s o1~\cite{openai2024learning} and DeepSeek-R1~\cite{guo2025deepseek} have achieved notable success in tasks such as mathematical problem solving by leveraging long chain-of-thought (CoT)~\cite{wei2022chain} reasoning. These models excel at strategies like mistake correction, step decomposition, and iterative refinement, resulting in more structured and extended reasoning. Recently, several studies~\cite{liu2025visual, huang2025vision} have adapted the DeepSeek-R1 framework to visual reasoning, training models to generate CoT-based outputs directly from visual inputs.
While these approaches focus on finetuning the reasoning model itself to perform end-to-end visual inference, our work takes an orthogonal perspective. Instead of modifying or retraining the reasoning model, we propose to train an autonomous agent that reasons about which tools to select to best assist a \emph{frozen} reasoning model in solving a given query. By learning to select supportive tools based on each query, our agent enhances performance without altering the model’s internal parameters (thereby preserving generalization to other tasks). This design also ensures broad compatibility across different visual reasoning models and offers a flexible, modular strategy for improving multimodal reasoning systems. ReTool~\cite{feng2025retool} is a concurrent work that uses RL to teach LLMs to invoke code execution tools for text-based reasoning. In contrast, our work tackles the challenging setting of visual reasoning with diverse tool choices, requiring agents to adaptively select the most effective tools.

\textbf{Visual Reasoning Tasks.} Some visual reasoning tasks, such as depth estimation and spatial reasoning~\cite{fu2024blink}, can be effectively solved using straightforward, specialized tools like depth estimators or object detectors. In contrast, we target more complex and cognitively demanding tasks, where optimal tool selection is query-dependent and may not be obvious.
Chart understanding~\cite{masry2022chartqa, lu2023mathvista} is a challenging task and a strong indicator of visual reasoning capabilities. It requires models to process numerical data, textual labels, and complex visual structures, not only identifying visual elements but also performing precise interpretation and measurement, such as accurately estimating bar heights in bar charts.
Geometry questions~\cite{lu2021inter, lu2023mathvista, zhang2024mathverse} pose a similarly demanding challenge, requiring fine-grained diagram understanding followed by text-based reasoning grounded in visual details. It evaluates the multi-modal logical reasoning abilities of VLMs by asking them to decode visual elements in diagrams, such as characters and geometric figures, and align them with conditions specified in textual questions for mathematical problem solving.
Our method achieves strong performance on both types of benchmarks, demonstrating the effectiveness of learning to select tools adaptively for chart and geometry reasoning tasks.

\section{Method}

In this section, we present VisTA, a reinforcement learning (RL) framework for tool-augmented visual reasoning. In contrast to previous methods that rely on training-free or fine-tuned strategies, VisTA empowers an agent to learn how to select tools through trial-and-error interaction, without requiring manual supervision. By harnessing the exploration-exploitation dynamics of RL, the agent adaptively chooses from a wide array of tools based on performance feedback. Notably, the core reasoning model is kept fixed, allowing the learned tool-selection policy to transfer across different reasoning backbones without the need for additional training. 

\begin{figure}[t]
\centering
\includegraphics[width=0.8\textwidth]{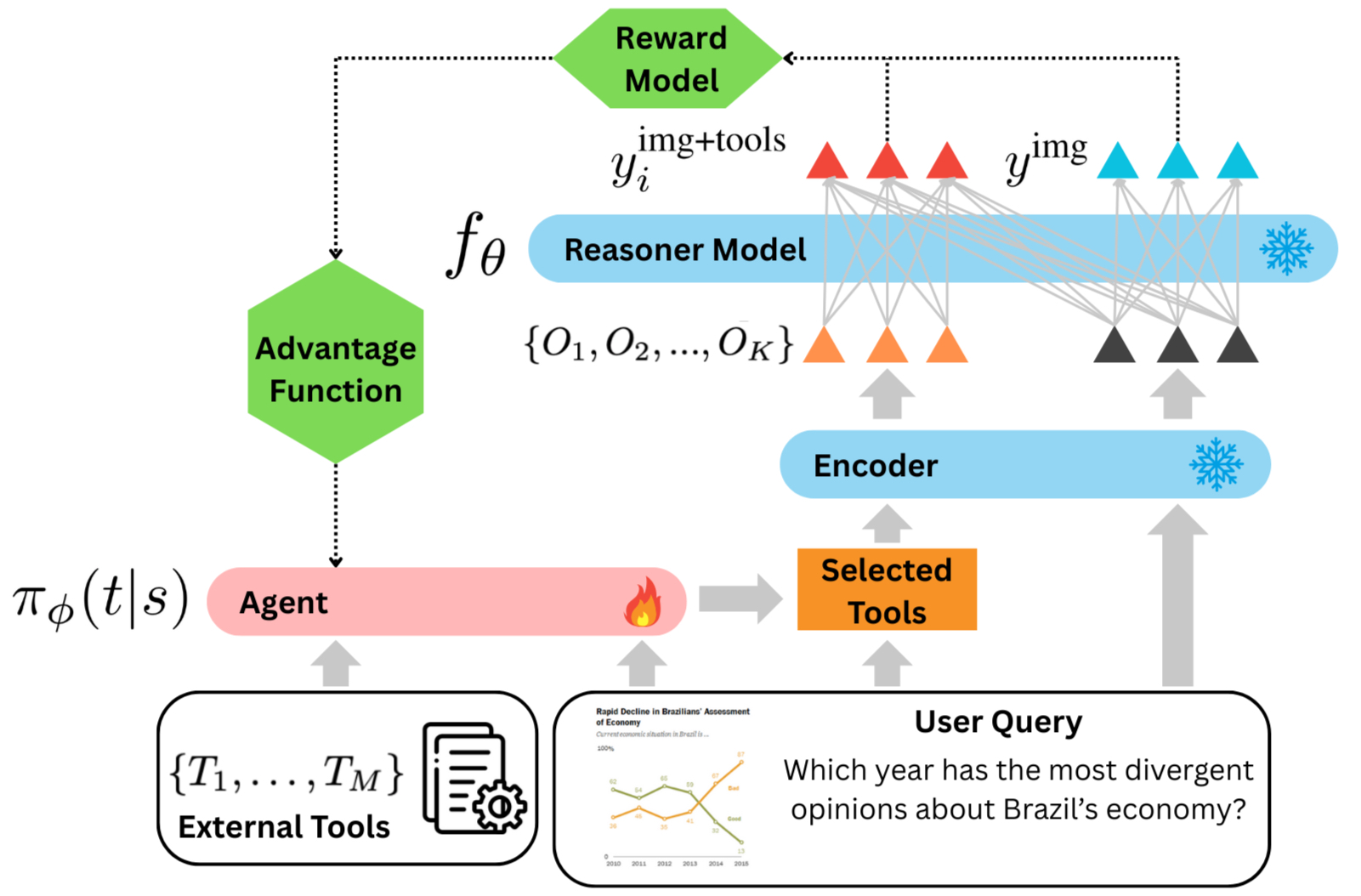}
\caption{\textbf{Policy Optimization.} Given a user query, the agent selects tools from a pre-defined set of external tools. The tools are applied to the image, and their outputs and the query are fed to a frozen reasoner model. Both the \emph{Direct Path} (query+image) and the \emph{Tool-Augmented Path} (query+tools+image) are evaluated to compute a reward signal, which is used to update the agent's tool-selection policy.}
\label{fig:main}
\end{figure}

\subsection{Problem Formulation}
Let $(q, I)$ denote a visual-language query, consisting of an image $I$ and an associated text query $q$, sampled from a task distribution $\mathcal{D}$. We consider a setting with a \textit{frozen} visual-language model (the \emph{reasoner}) $f_{\theta}$ and a library of external tools $\mathcal{T} = \{T_1, \ldots, T_M\}$, where $M$ is the total number of available tools. These tools include diverse specialized vision modules; e.g., in chart understanding tasks, we utilize chart-to-SVG converters, chart-to-table extractors, and chart-to-caption generators.

We define the agent's observation or \emph{state} as \( s = (q, I) \), encompassing both the image and its corresponding query. Our objective is to learn a \emph{selection policy} \( \pi_{\phi}(t \mid s) \), implemented as a vision-language model (the \emph{agent}), that determines which tools to deploy for a given query. Formally, the selection policy maps the state \( s \) to a sequence of selected tools \( t = \langle T^{(1)}, \ldots, T^{(K)} \rangle \), where \( T^{(i)} \in \mathcal{T} \) and \( K \leq M \). Here, \( K \) represents the number of tools selected for the specific query, which adaptively varies based on task complexity.

\subsection{VisTA Framework}
Our VisTA framework implements an end-to-end pipeline to learn a policy for dynamic tool selection in visual reasoning tasks (Figure~\ref{fig:main}), leveraging reinforcement learning to enable systematic exploration of tool combinations. The pipeline operates as follows:

\begin{enumerate}[leftmargin=*, nosep]
    \item The vision-language agent observes the state \( s = (q, I) \) and selects a sequence of tools \( t = \langle T^{(1)}, \ldots, \\ T^{(K)} \rangle \) via a policy \( \pi_{\phi}(t \mid s) \).
    
    \item Each selected tool is executed on the image to produce outputs \( o^{(i)} = T^{(i)}(I) \) for \( i \in \{1, \ldots, K\} \).
    
    \item These outputs are combined with the original inputs to form an augmented prompt.
    
    \item The frozen reasoner \( f_{\theta} \) processes the augmented prompt to produce the final answer \( y^{\text{img+tools}} = f_{\theta}(q, I, \{o^{(1)}, \ldots, o^{(K)}\}) \).
\end{enumerate}

During training, we also compute a baseline prediction \( y^{\text{img}} = f_{\theta}(q, I) \) using only the original query and image, which enables us to measure the impact of selected tools on reasoning performance, as will be detailed in our reward formulation in the following section.

\subsection{Policy Optimization}

To train the agent to discover effective tool combinations through systematic exploration, we build upon the GRPO algorithm~\cite{shao2024deepseekmath}. The key difference is that, to adapt it for visual tool selection, we design a novel task-specific reward function.

Given a context, a particular tool may be helpful, harmful, or neutral from the reasoner's perspective. We reward the agent if the selected tool enables the reasoner to answer the question correctly. If the tool does not lead to a correct answer, the reward depends on the reasoner's performance without the tool-augmented input. Specifically, if the reasoner also fails without the tool, the reward is set to 0, since the tool had no effect on the outcome. However, if the reasoner could have answered correctly on its own, we assign a reward of -0.5, penalizing the negative influence on the final answer.

More formally, at each training step, given an input state \( s = (q, I) \), we sample \( G \) tool selection candidates \( \{t_j\}_{j=1}^G \) from the current policy \( \pi_{\phi} \). For each candidate \( t_j \), we compute a tool-aware reward \( r_j \) by comparing the reasoner’s predictions with and without tools:
\[
r_j =
\begin{cases}
+1 & \text{if } y^\text{img} \neq y^\ast \text{ and } y^\text{img+tools}_j = y^\ast \quad (\text{tools help}) \\
-0.5 & \text{if } y^\text{img} = y^\ast \text{ and } y^\text{img+tools}_j \neq y^\ast \quad (\text{tools hurt}) \\
0 & \text{if } y^\text{img} \neq y^\ast \text{ and } y^\text{img+tools}_j \neq y^\ast \quad (\text{no change}) \\
+1 & \text{if } y^\text{img} = y^\ast \text{ and } y^\text{img+tools}_j = y^\ast \quad (\text{tools neutral})
\end{cases}
\]
where \( y^\text{img} \) is the prediction from the reasoner using only the question, \( y^\text{img+tools}_j \) is the prediction with selected tools from \( t_j \), and \( y^\ast \) is the ground truth. This reward design incentivizes tool selections that lead to correct predictions when the image alone is insufficient, while penalizing those that hurt performance. We then calculate each candidate’s group-relative advantage as \( A_j = \frac{r_j - \text{mean}(\{r_1, \dots, r_G\})}{\text{std}(\{r_1, \dots, r_G\})} \). This normalization contextualizes each tool selection's performance relative to other candidates, enabling the agent to distinguish which tool combinations performed better for a specific query regardless of absolute reward values. Finally, the policy \( \pi_{\phi} \) is updated by maximizing the following:
\begin{equation*}
\mathcal{J}_{\text{VisTA}}(\phi) = 
\mathbb{E}_{s \sim \mathcal{D}} \mathbb{E}_{\{t_j\}_{j=1}^G \sim \pi_{\phi_\text{old}}(\cdot|s)} \left[
\frac{1}{G} \sum_{j=1}^G 
    \min\left(r_j^\text{ratio} A_j, \text{clip}(r_j^\text{ratio}, 1 \pm \epsilon) A_j \right)
- \beta \mathbb{D}_{\text{KL}}(\pi_\phi \| \pi_\text{ref})
\right]
\end{equation*}

where \( r_j^\text{ratio} = \frac{\pi_{\phi}(t_j|s)}{\pi_{\phi_\text{old}}(t_j|s)} \), \( \pi_{\phi_\text{old}} \) is the previous policy, and \( \pi_\text{ref} \) is a fixed reference policy.
This objective balances exploration and exploitation through ratio clipping and KL regularization. The ratio term prevents large policy updates, while the KL divergence term maintains stability by keeping the policy close to a reference distribution. Through this optimization, tools that consistently improve reasoning performance receive higher selection probabilities, allowing the agent to discover effective tool combinations based on empirical evidence rather than predefined rules.

\subsection{Tool Selection Prompting}
We design a structured prompt template to guide the agent's tool selection process. For chart understanding tasks, this template presents 9 tools (indexed 0-8) organized across three functional categories:
{\small\ttfamily ``You are an expert agent tasked with selecting tools to solve chart reasoning tasks. You have access to 9 tools indexed from 0 to 8, each belonging to one of three functional types: type1, type2, \\and type3...
Function: 0: type1 (A), 1: type1 (B), 2: type1 (C), 3: type2 (D), 4: type2 \\(E) ...
Given a chart and a query \texttt{\{Question\}}, select the index number(s) of tools that are most helpful. Output only selected indices as a comma-separated list within \texttt{<answer>} tags."}

\noindent This prompt structure provides only minimal tool categorization without detailed function descriptions or usage demonstrations that training-free methods typically rely on. Instead, our framework incentivizes the agent to empirically discover and adaptively select the most effective tools through reinforcement learning. This approach is particularly valuable when integrating tools with unknown, partially documented, or inconsistently performing capabilities.

\section{Experiments}

We evaluate our VisTA framework on visual reasoning benchmarks, comparing to training-free baselines, alternative RL-based methods. We also analyze VisTA's tool selection strategies and distribution, and agent behavior dynamics over training. 

\subsection{Experimental Setup}

\begin{table}[t]
\scriptsize
\centering
\begin{tabular}{l c c c c c}
\toprule
Method & Agent Model & Reasoning Model &  ChartQA & ChartQA(OoD)  & Geometry3K\\
\midrule
Training-Free & - & QwenVL 7B &76.4 & 62.3 & 54.0\\
Training-Free & QwenVL 7B & QwenVL 7B & 76.1 &  66.8 & 51.3\\
Training-Free & GPT4o & QwenVL 7B & 73.0 &  66.4 & 51.5 \\
RL            & -     & QwenVL 7B & 77.5 &  64.3 & 41.0 \\
\rowcolor[HTML]{cef8d1} Ours & QwenVL 7B & QwenVL 7B & 79.4 & 73.2  & 55.6\\
\midrule
Training-Free & - & GPT4o & 84.3 & 50.1  & 50.1\\
Training-Free & QwenVL 7B & GPT4o & 82.3 & 67.1  & 48.7\\
Training-Free & GPT4o & GPT4o & 84.6 & 73.3  & 49.5\\
\rowcolor[HTML]{cef8d1} Ours & QwenVL 7B & GPT4o & 88.9 & 76.8  & 52.4\\
\bottomrule
\end{tabular}
\vspace{1pt}
\caption{\textbf{Main Results.} These results highlight VisTA’s ability to support complex, multi-modal reasoning where tools provide complementary visual understanding, and its flexibility and compatibility with stronger reasoning models like GPT-4o at deployment time.}
\label{table:main_result}
\vspace{-0.15in}
\end{table}

We conduct experiments on two main datasets: ChartQA~\cite{masry2022chartqa} and Geometry3K~\cite{lu2021inter}. Both datasets provide paired visual inputs and questions, but differ in their reasoning styles and evaluation protocols. We train our VisTA agent using the training sets of both datasets, and evaluate on their respective test sets. For ChartQA, in addition to the original test set, we construct an out-of-distribution (OoD) variant by manually removing textual labels from the charts. This setting is designed to assess the model's reliance on textual cues versus its ability to reason visually. As a result, we evaluate on three test sets in total: ChartQA (in-domain), ChartQA-OoD (perturbed version), and Geometry3K. ChartQA adopts an open-ended answer format, so we report relaxed accuracy (within 5\% of the gold answer) as the evaluation metric, which credits semantically correct answers. In contrast, Geometry3K is a multiple-choice benchmark and uses standard accuracy to measure exact match with the correct option.

For both ChartQA and Geometry3K, we randomly sample ~800 examples for training and evaluate on their full test sets. We train our method using 8 NVIDIA A100 GPU, with a batch size of 8 and 4 answer generations per query to encourage diverse tool interactions. We use the AdamW\cite{loshchilov2017decoupled} optimizer with a learning rate of $5 \times 10^{-5}$. The agent is trained for 100 iterations. During inference, we use only the Tool-Augmented prediction path. 

We curate benchmark-specific tool pools tailored to the reasoning demands of each domain. For ChartQA, we construct a pool of 9 tools grouped into three functional types: chart-to-SVG, chart-to-table, and chart-to-caption, with 3 tools per type. chart-to-table tools (T0: UniChart~\cite{masry2023unichart}, T1: DePlot~\cite{liu2022deplot}, T2: ChartMoE~\cite{xu2024chartmoe}) convert charts into structured format tables, chart-to-SVG tools (T3: OpenCV~\cite{bradski2000opencv}, T4: ChartDet~\cite{yan2023context}, T5: ChartQCR~\cite{luo2021chartocr}) extract geometry elements (bar, line, pie), and color from charts, and captioning modules (T6: ChartAssistant~\cite{meng2024chartassisstant}, T7: ChartVLM~\cite{xia2024chartx}, T8: QwenVL-32B~\cite{bai2025qwen2}) summarize visual content at a global level. For Geometry3K, we use a smaller pool of 4 tools categorized into two types: symbolic parsers and specialist geometry solvers, with 2 tools in each category. Symbolic parsers extract entities and relations (e.g., angles, labels), while the specialist models take in the question and diagram to generate a sequence of steps which to solve the problem. For the symbolic parser, we use T0: DiagramFormalizer \cite{zhang2025diagram} and T1: Inter-GPS \cite{lu2021inter}. For our specialist solvers, we use T2: G-LLaVA 13B \cite{gao2023g} and T3: MultiMath \cite{peng2024multimath}. Each tool within a type is implemented independently, allowing the agent to learn fine-grained preferences among tools with overlapping but distinct capabilities.

\subsection{Main Results}

We first evaluate our VisTA framework across three visual reasoning benchmarks—ChartQA, ChartQA-OoD, and Geometry3K, comparing against training-free baselines,  alternative RL-based approaches, and state-of-the-art VLMs~\cite{team2024gemini, anthropic2024claude, beyer2024paligemma, abdin2024phi, internvl2, hurst2024gpt, llama3, tong2024cambrian, agrawal2024pixtral, deitke2024molmo}. 

Table~\ref{table:main_result} summarizes the results. For tool selection, we use QwenVL-7B~\cite{bai2025qwen2} as the agent and keep the reasoning model (QwenVL-7B or GPT4o) frozen during training.  When both the agent and reasoner use QwenVL 7B, our method consistently outperforms baselines across all benchmarks. On ChartQA, VisTA achieves 79.4\% accuracy, improving over the best training-free baseline (76.4\%) by 3.0 points. On Geometry3K benchmark, which requires fine-grained diagram understanding and multi-step symbolic reasoning, VisTA achieves 55.6\%, outperforming the best training-free baseline (54.0\%). This result highlights VisTA’s ability to support complex, multi-modal reasoning where tools provide complementary visual understanding.

A key strength of our method is its transferability. Without any retraining, the tool-selection policy learned with QwenVL 7B can be paired with GPT-4o as the reasoner. 

\begin{table}[t!]
\centering
\footnotesize
\setlength{\tabcolsep}{2pt}
\begin{tabular}{l c c }
\toprule
Method                                  & ChartQA & Geometry3K \\
\midrule
\rowcolor[HTML]{cef8d1}
Ours                                    & 88.9    & 55.6      \\
\midrule
Intern2VL-8B~\cite{internvl2}           & 83.3    & 26.5       \\
Intern2VL-8B-ShortCoT~\cite{internvl2}  & –       & 29.7       \\
Geo-Intern2VL-8B~\cite{internvl2}       & –       & 30.7       \\
G-LLAVA-7B~\cite{gao2023g}              & –       & 22.4       \\
Math-LLAVA-13B~\cite{shi2024math}       & –       & 33.1       \\
QvQ-72B-Preview                         & –       & 29.4       \\
RedStar-Geo-8B~\cite{xu2025redstar}     & –       & 33.6       \\
\midrule
GPT4v~\cite{achiam2023gpt}              & 78.1    & –          \\
GPT4o-0513~\cite{hurst2024gpt}          & 85.7    & –          \\
Gemini 1.5 Flash~\cite{team2024gemini}  & 85.4    & –          \\
Gemini 1.5 Pro~\cite{team2024gemini}    & 87.2    & –          \\
Claude-3 Haiku~\cite{anthropic2024claude} & 81.7 & –          \\
Claude-3 Opus~\cite{anthropic2024claude}  & 80.8 & –          \\
Claude-3.5 Sonnet~\cite{anthropic2024claude} & 90.8 & –      \\
PaliGemma-mix-3B~\cite{beyer2024paligemma}  & 33.7 & –         \\
Phi3.5-Vision-4B~\cite{abdin2024phi}       & 81.8 & –         \\
InternVL2-Llama-3-76B~\cite{internvl2}     & 88.4 & –         \\
Pixtral-12B~\cite{agrawal2024pixtral}      & 81.8 & –         \\
Llama-3.2V-11B-Instruct~\cite{llama3}      & 83.4 & –         \\
Llama-3.2V-90B-Instruct~\cite{llama3}      & 85.5 & –         \\
LLaVA-1.5-7B~\cite{liu2024improved}        & 17.8 & –         \\
LLaVA-1.5-13B~\cite{liu2024improved}       & 18.2 & –         \\
xGen-MM-interleave-4B~\cite{xue2024xgen}   & 60.0 & –         \\
Cambrian-1-8B~\cite{tong2024cambrian}      & 73.3 & –         \\
Cambrian-1-34B~\cite{tong2024cambrian}     & 75.6 & –         \\
LLaVA OneVision-7B~\cite{li2024llava}      & 80.0 & –         \\
LLaVA OneVision-72B~\cite{li2024llava}     & 83.7 & –         \\
MolmoE-1B~\cite{deitke2024molmo}           & 78.0 & –         \\
Molmo-7B-O~\cite{deitke2024molmo}          & 80.4 & –         \\
Molmo-7B-D~\cite{deitke2024molmo}          & 84.1 & –         \\
Molmo-72B~\cite{deitke2024molmo}           & 87.3 & –         \\
\bottomrule
\end{tabular}
\caption{\textbf{Comparison to state-of-the-art VLMs.}  For our approach, the agent model is QwenVL 7B, and the reasoners are GPT4o for ChartQA, and QwenVL 7B for Geometry3K.}
\label{table:sota}
\end{table}

In this setting, we achieve 88.9\% on ChartQA and 76.8\% on ChartQA-OoD, surpassing the best training-free GPT-4o baseline by 3.5 points on both, and also achieve 52.4\% on Geometry3K, outperforming the best baseline (49.5\%) by 2.9 points. 
This demonstrates VisTA’s flexibility and compatibility with stronger reasoning models at deployment time.  We also compare against an RL baseline where the reasoner is directly trained using GRPO~\cite{guo2025deepseek} to generate chain-of-thought reasoning in a \texttt{<think>} block followed by a final \texttt{<answer>} block, without using tools. VisTA outperforms this RL-trained reasoner by 1.9 points on ChartQA and a substantial 8.9 points on ChartQA-OoD, showing that tool-augmented reasoning offers greater gains than direct model optimization.

The advantage of our method over the baselines is even more pronounced on ChartQA-OoD, where VisTA reaches 73.2\%, a 6.4-point gain over the best baseline (66.8\%).  This result demonstrates that by learning to select visual tools via RL, our VisTA agent framework is able to reason better visually than other baselines, which may rely more on textual cues.

Finally, Table~\ref{table:sota} shows the comparison to state-of-the-art VLMs. VisTA achieves the best performance on Geometry3K, significantly outperforming all prior methods. On ChartQA, VisTA ranks second overall, only slightly behind Claude-3.5 Sonnet (90.8 vs. 88.9), and surpasses other strong baselines such as Molmo-72B, Gemini 1.5 Pro, and InternVL2-Llama-3. This demonstrates that our approach is both highly effective on complex chart reasoning tasks and substantially more capable on geometric benchmarks.

\subsection{Tool Selection Analysis}

\begin{figure}
\centering
\includegraphics[width=0.6\linewidth]{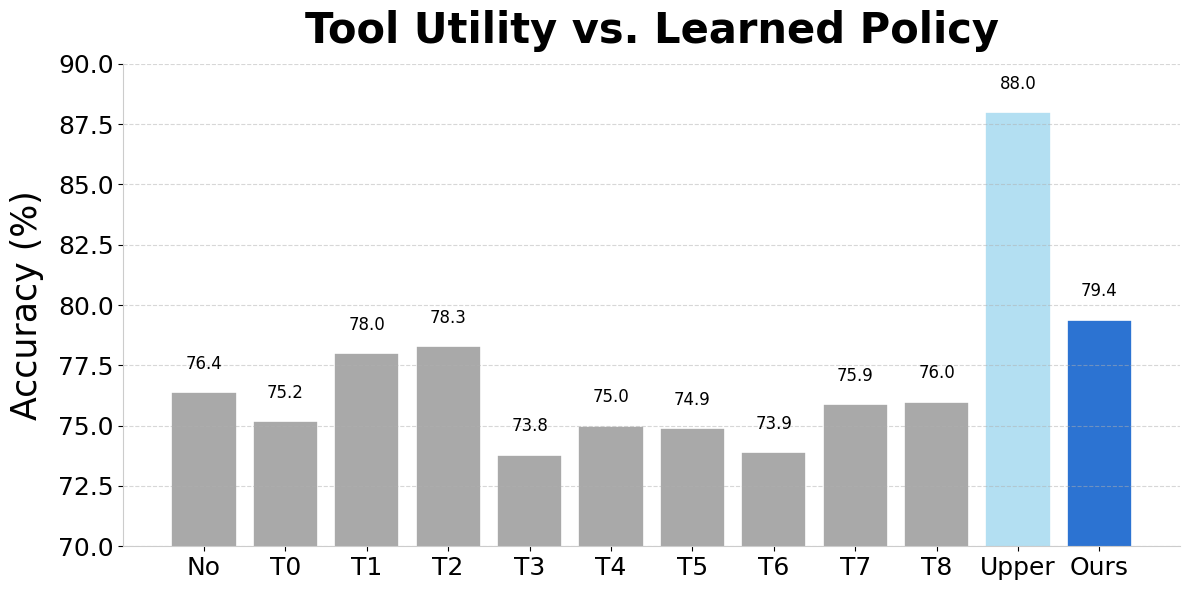}  
    \caption{Comparison of ChartQA accuracy across individual tools (T0–T8), the no-tool baseline (No), our RL-based selection policy (Ours), and a pseudo-upper bound (Upper).}
    \label{fig:agent_behavior}
\end{figure}

To evaluate whether our tool selection policy effectively learns to combine and select the most appropriate tools for each query, we compare its performance against using individual tools in isolation. Specifically, we feed each tool’s output along with the original input to the frozen reasoner and record its accuracy on the ChartQA benchmark.  We also compute a pseudo-upper bound of 88.0\% by treating a query as correct if any single tool enables the reasoner to produce the correct answer. This serves as a loose upper limit on what could be achieved with perfect single-tool selection, though it does not account for the benefits of combining multiple tools. 

Fig.~\ref{fig:agent_behavior} shows the performance of each tool individually (T0–T8), as well as the no-tool baseline (76.4\%). While certain tools, such as T2 (78.3\%) and T1 (78.0\%), improve upon the no-tool baseline, the large gap to the pseudo-upper bound (88.0\%) suggests that no single tool consistently performs best across all queries. Different tools appear to be optimal for different subsets of the data.  Ideally, a well-trained policy should learn to select the most effective tool(s) for each specific query, achieving performance that surpasses any static tool choice. Our method achieves 79.4\% accuracy, outperforming all individual tools. This suggests that the policy learns to go beyond fixed tool usage and adapt its selection based on query-specific context.  Our policy also closes some gap between the best individual tool and the pseudo-upper bound, indicating progress toward optimal tool selection without explicit supervision.

\subsection{Agent Behavior Evolution During RL Training}

To analyze whether the agent is learning to prefer more effective tools, we track the correlation between tool usage frequency and individual tool performance over training. Specifically, every 10 iterations, we compute the Pearson correlation coefficient between the usage counts of each tool and their corresponding standalone accuracy (as reported in Fig.~\ref{fig:agent_behavior}).

\begin{figure}
\centering
\includegraphics[width=0.6\linewidth]{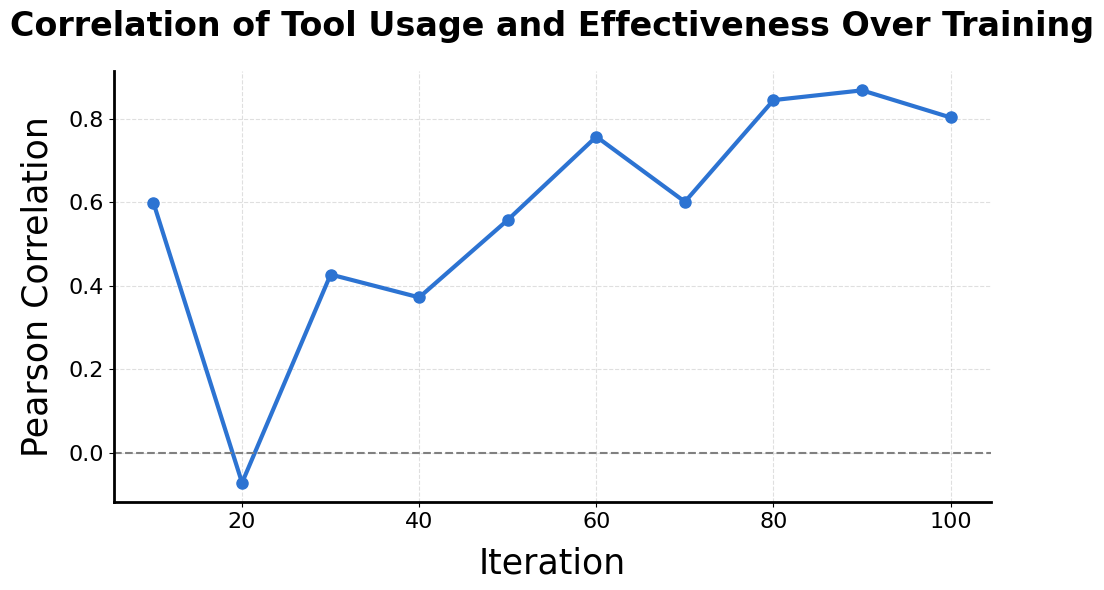}  
    \caption{Pearson correlation between tool usage frequency and individual tool performance.}
    \label{fig:tool_correlation}
\end{figure}

Figure~\ref{fig:tool_correlation} shows the evolution of this correlation across training iterations. Despite some initial fluctuations, we observe a clear upward trend, with the correlation increasing from near zero to over 0.8 as training progresses. This indicates that the agent is gradually aligning its tool selection strategy with the relative utility of each tool—favoring those that contribute more to the reasoner's accuracy.  These results suggest that our RL-based policy does not rely on fixed heuristics but instead learns to discriminate among tools based on their empirical contribution to task success. The emergence of this alignment over time provides further evidence that the agent is effectively adapting its behavior through reinforcement feedback.

\subsection{Tools Selection Distribution}

\begin{figure}[t]
    \centering
    \includegraphics[width=0.7\textwidth]{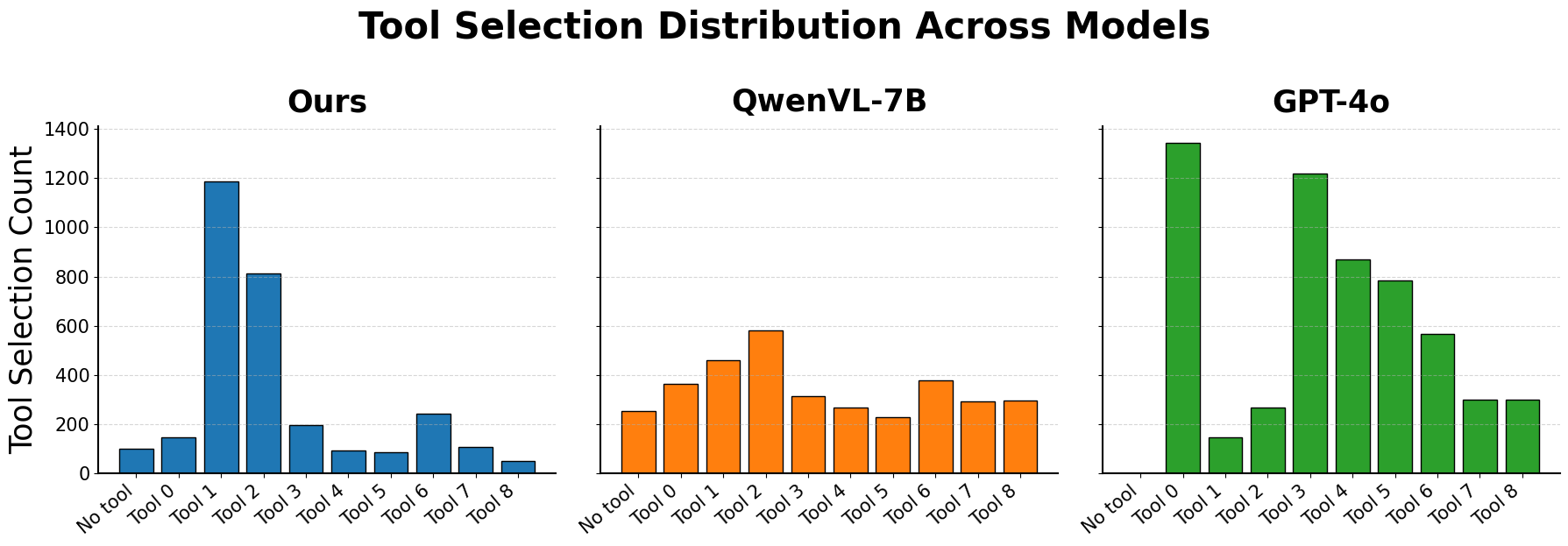}
    \caption{Tool selection frequency across our RL-trained agent, QwenVL-7B, and GPT-4o. Our method strongly favors effective tools (Tools 1 and 2) and avoids less useful ones, while QwenVL-7B shows a uniform distribution and GPT-4o selects broadly without clear alignment to tool performance.}
    \label{fig:tool_distribution}
\end{figure}

Figure~\ref{fig:tool_distribution} illustrates the tool selection distribution on the test set for our RL-trained agent, as well as for the training-free baselines using QwenVL-7B and GPT-4o. Our method clearly shows a strong preference for Tool 1 and Tool 2, both are chart-to-table tools, which are among the most effective tools based on individual performance (see Fig.~\ref{fig:agent_behavior}). In contrast, low-performing tools such as Tool 3 (chart-to-SVG) and Tool 6 (caption module), are selected far less frequently, suggesting that the learned policy has effectively adapted to favor high-utility tools based on empirical feedback.

The QwenVL-7B baseline, which operates in a training-free manner without reinforcement feedback, exhibits a more balanced selection pattern that resembles a near-normal distribution. This indicates that it lacks strong preferences and does not consistently prioritize the most effective tools. Meanwhile, GPT-4o tends to select more tools per query, rarely opts for no tool, and distributes its selections across a broader set of tools. However, this broader usage still lacks clear alignment with tool effectiveness, showing no strong correlation between selection frequency and tool performance.

These differences highlight the benefit of learning a tool selection strategy through RL. Unlike training-free approaches that rely solely on static prompt understanding, our agent adapts its behavior based on downstream task outcomes, leading to more effective tool use.

\subsection{Comparison with Random and All Selection Strategies}

To further evaluate the effectiveness of our learned tool selection policy, we compare it with two alternative tool selection strategies:
All Tools: In this baseline, all nine tool outputs are simultaneously provided to the reasoning model alongside the original image, creating a comprehensive but potentially overwhelming input.
Random Tools: This baseline randomly selects a fixed number of tools for each query. Since our policy selects approximately 1.2 tools per query on average, we randomly sample 2 tools per query for a fair comparison.

Table~\ref{tab:ablation_random_all} presents the comparative results across different reasoning models (QwenVL-7B and GPT-4o). Our selective approach consistently outperforms both baselines, demonstrating that strategic tool selection provides substantial benefits over both comprehensive and random approaches.

\begin{table}[h]
\centering
\scriptsize
\begin{tabular}{lccc}
\toprule
Reasoning Model & All Tools & Random Tools & Ours \\
\midrule
QwenVL-7B & 74.6 & 75.8 & \textbf{79.4} \\
GPT-4o & 81.3 & 83.5 & \textbf{88.1} \\
\bottomrule
\end{tabular}
\caption{Performance comparison of different tool selection strategies.}
\label{tab:ablation_random_all}
\end{table}

The performance gap between our method and the All Tools baseline (4.8 and 6.8 percentage points for QwenVL-7B and GPT-4o, respectively) reveals that simply providing more tool outputs can be counterproductive. This suggests that extraneous information may confuse the reasoning model, highlighting the importance of selective augmentation. Similarly, our method outperforms the Random Tools baseline by 3.6 and 4.6 points, highlighting that effective reasoning depends on tool choice, not just quantity.

\subsection{Evaluation on MathVerse Geometry Benchmark}
We evaluate our method on the MathVerse~\cite{zhang2024mathverse} benchmark, a recent dataset designed to assess visual reasoning in geometry problems. Since our toolset focuses on 2D plane geometry, we report results on the corresponding subset of MathVerse in Table~\ref{tab:mathverse}, which includes 2,550 samples out of the full 3,940.

Importantly, we do not retrain the agent on MathVerse. Instead, we directly apply the agent trained on Geometry3K~\cite{lu2021inter} to assess its zero-shot generalization capability to a new benchmark. As shown in Table~\ref{tab:mathverse}, our method outperforms all training-free baselines, including setups where GPT-4o is directly prompted with tool outputs. 

Additionally, we compare our method with several recent state-of-the-art (SOTA) VLMs, including GPT-o1, Claude-3.5 Sonnet, and Gemini-2.0-Flash. Our method achieves the best performance across all configurations, including a strong result of 73.1 when combined with GPT-o1 as the reasoner—surpassing the best prior SOTA accuracy of 67.7. These results demonstrate the robustness and generalizability of our tool selection policy, even when applied to more structurally diverse and challenging geometry benchmarks.

\begin{table}[h]
\footnotesize
\centering
\begin{tabular}{l c c c }
\toprule
Method & Agent Model & Reasoning Model &  MathVerse \\
\midrule
Training-Free & - & GPT4o & 50.2 \\
Training-Free & QwenVL 7B & GPT4o & 51.4 \\
Training-Free & GPT4o & GPT4o & 53.6 \\
\rowcolor[HTML]{cef8d1} Ours & QwenVL 7B & GPT4o & 56.1 \\
\midrule
Training-Free & - & Claude-3.5-Sonnet & 58.2\\
Training-Free & - & Gemini-2.0-Flash & 65.6\\
Training-Free & - & GPT-o1 & 67.7\\
\rowcolor[HTML]{cef8d1} Ours & QwenVL 7B & GPT-o1 & 73.1\\
\bottomrule
\end{tabular}
\caption{Results on the plane geometry subset of the MathVerse benchmark. Our method achieves the best performance, demonstrating its effectiveness on structurally diverse geometry reasoning tasks.}
\label{tab:mathverse}
\end{table}
\subsection{Results on BlindTest Benchmark}

Finally, to assess the generalizability of our framework, we evaluate on  BlindTest~\cite{rahmanzadehgervi2024vision}, which targets low-level vision tasks requiring fine-grained spatial understanding where even state-of-the-art VLMs like GPT-4o~\cite{hurst2024gpt} and Gemini 1.5 Pro~\cite{team2024gemini} underperform. Following our previous training setup, we train a QwenVL-7B agent to select from six tools, grouped into three categories, with each category responsible for detecting a specific geometric element: lines, circles, or rectangles. Each category includes two tools to offer complementary perspectives. As shown in Table~\ref{tab:blindtest}, our method achieves higher accuracy than both training-free baselines, highlighting the advantage of learned tool selection in scenarios that demand fine-grained visual reasoning. 

\begin{table}[h]
\centering
\scriptsize
\begin{tabular}{lccc}
\toprule
Method & Agent Model & Reasoning Model & BlindTest \\
\midrule
Training-Free & - & GPT4o & 48.5 \\
Training-Free & GPT4o & GPT4o & 51.8 \\
\rowcolor[HTML]{cef8d1} Ours & QwenVL 7B & GPT4o & \textbf{53.4} \\
\bottomrule
\end{tabular}
\caption{Performance of different tool selection strategies on BlindTest.}
\label{tab:blindtest}
\end{table}

\subsection{Comparison with Visual Sketchpad on Geometry3K}

To further benchmark our method, we compare against Visual Sketchpad~\cite{hu2024visual}, a recent state-of-the-art system on Geometry3K. Visual Sketchpad enables VLMs to draw with lines, boxes, marks, which better facilitates visual perception and reasoning. For a fair comparison, we follow their evaluation protocol and test our method on the same curated subset of Geometry3K\textsuperscript{*}, which is smaller than the official test set. As shown in Table\ref{tab:sketchpad}, our method still outperforms Visual Sketchpad. This result highlights the strength of our learned tool selection policy in driving effective visual reasoning.

\begin{table}[h]
\footnotesize
\centering
\begin{tabular}{l c c c }
\toprule
Method & Agent Model & Reasoning Model &  Geometry3K\textsuperscript{*} \\
\midrule

Visual Sketchpad~\cite{hu2024visual} & - & GPT4o & 66.7 \\
\rowcolor[HTML]{cef8d1} Ours & QwenVL 7B & GPT4o & \textbf{68.8} \\
\bottomrule
\end{tabular}
\caption{Comparison with Visual Sketchpad on a Geometry3K\textsuperscript{*} test subset. We follow their evaluation protocol using the same curated set. \textsuperscript{*}Indicates the smaller subset used by Visual Sketchpad. Our method achieves higher accuracy, demonstrating the strength of our tool selection policy.}
\label{tab:sketchpad}
\end{table}

\subsection{Visualizations}\label{app:viz}
In this section, we provide qualitative examples showing how our agent selects appropriate tools according to the query. 

\begin{figure}[t]
    \centering
    \includegraphics[width=0.5\textwidth]{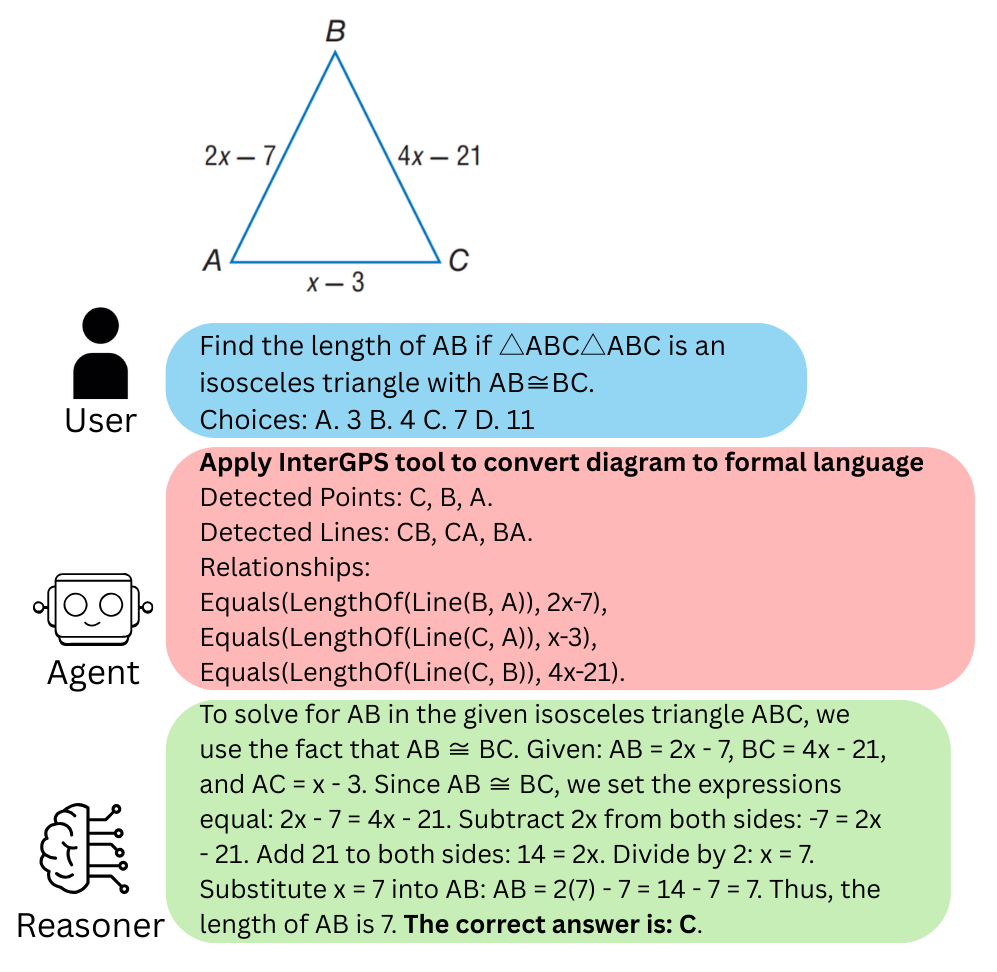}
    \caption{\textbf{Tool Selection for a Geometry Question}. Our agent selects the formal diagram parser, Inter-GPS, which accurately extracts essential details from the diagram and represents them in a formal structure. Leveraging this representation, the reasoner is able to determine the correct answer.}
    \label{fig:geom_viz}
\end{figure} 

From the Geometry3k dataset, we present an example in Figure \ref{fig:geom_viz}. Based on the context, our agent selects the Inter-GPS tool, which represents the objects in the diagram and their relationships using formal language. This structured information is then used by the Reasoner model to correctly answer the question.

\begin{figure}[t]
    \centering
    \includegraphics[width=0.8\textwidth]{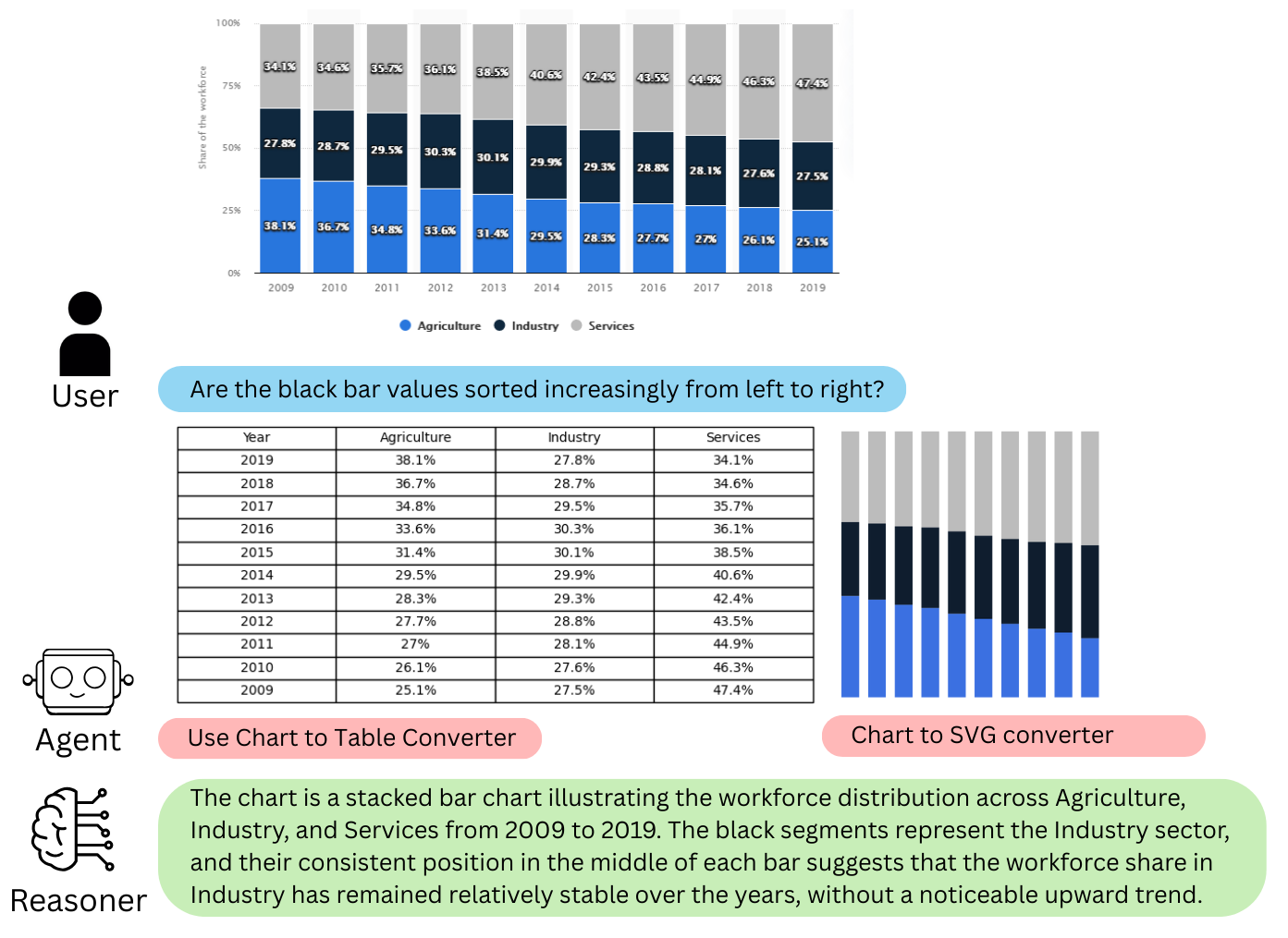}
    \caption{\textbf{Tool Selection for a Chart Question}. An example demonstrating a case where the agent calls a set of complementary tools. This question requires understanding both numeric values (extracted from the table) and color information (extracted from the SVG). The reasoner combines this information to arrive at the correct answer.}
    \label{fig:chart_viz}
\end{figure} 

Figure \ref{fig:chart_viz} presents a qualitative example of chart understanding. The question is complex, requiring reasoning about the sizes of various objects identified by specific colors. To handle this, our agent invokes both a chart-to-table converter to extract numeric values (shown as an image for illustration) and an image-to-SVG converter to capture color information. Using this combined input, the reasoner is able to correctly answer the question.

\subsection{Constructing the ChartQA-OoD Test Set}
Chart-based question answering poses unique challenges, requiring models to reason over numerical values, textual labels, and complex visual structures. Effective chart comprehension demands not only the identification of visual elements but also precise interpretation—such as accurately estimating bar heights in bar charts.

To better evaluate the robustness and reasoning capabilities of VLMs, we construct an out-of-distribution (OoD) test set for ChartQA~\cite{masry2022chartqa} with two targeted perturbations. We manually remove textual labels from geometric elements (e.g., bars and points) in each chart. This isolates the model’s reliance on visual features by eliminating access to direct numeric answers via OCR. To further test the models' spatial reasoning robustness, we apply random geometric perturbations to the charts. With a 50\% chance, each chart is either horizontally or vertically stretched to twice its original width or height. These transformations maintain semantic structure but introduce visual variability. The resulting ChartQA-OoD test set allows us to probe whether models depend heavily on textual cues and how sensitive their reasoning is to mild visual distortions. This setup provides a more rigorous benchmark for evaluating the generalization and visual understanding capabilities of VLMs.

\section{Discussion and Conclusion}

We introduced VisTA, an RL framework enabling visual agents to autonomously select effective external tools for multimodal reasoning. Unlike prior methods, VisTA learns adaptive tool-selection strategies without explicit supervision. Our experiments showed significant accuracy gains over strong baselines, highlighting VisTA's potential for robust, flexible visual reasoning.

\textbf{Limitations.} Our framework enables agents to learn visual tool selection through experience, but it currently does not handle cases requiring sequential composition of multiple tools. Exploring this sequential tool-composition capability represents a promising direction for future research. Second, our framework currently relies on a fixed, manually curated set of tools, which could be enhanced by methods for automatically discovering and integrating new tools. Developing such automated tool-selection and integration methods could significantly improve the scalability and adaptability of our approach.

\textbf{Broader Impact.} Our framework enhances the ability of AI systems to autonomously learn when and how to utilize external tools, such as image captioners and object detectors, potentially improving generalization and robustness in multimodal environments. While our current study focuses on relatively low-stakes settings, such as chart understanding and geometry problems, deploying similar approaches in high-stakes domains like healthcare could introduce subtle yet significant errors due to incorrect tool usage. Ensuring reliability, transparency, and appropriate human oversight in these sensitive applications will be crucial.


\bibliographystyle{unsrt}
\bibliography{ref}



\end{document}